# Order Effects for Queries in Intelligent Systems

Subhash Kak

**Abstract.**
This paper examines common assumptions regarding the decision-making internal environment for intelligent agents and investigates issues related to processing of memory and belief states to help obtain better understanding of the responses. In specific, we consider order effects and discuss both classical and non-classical explanations for them. We also consider implicit cognition and explore if certain inaccessible states may be best modeled as quantum states. We propose that the hypothesis that quantum states are at the basis of order effects be tested on large databases such as those related to medical treatment and drug efficacy. A problem involving a maze network is considered and comparisons made between classical and quantum decision scenarios for it.

Keywords: reasoning systems, order effects, maze networks, quantum cognition

**Introduction**

The AI researcher is interested in developing systems that make optimal (or best) choices under conditions of uncertainty and specified limitations on computing resources. Since engineered and natural systems interact (as in health-care systems or human-machine interaction), there exist related problems of the logic underlying the decisions of human agents [1]. If the human decisions do not appear to be optimal, is that because the human actors are not rational, or are the assumptions regarding the decision-making internal environment different from what is assumed? One unique aspects of the behavior of intelligent agents arises from the fact that cognition has autonomous processes generally subsumed under the heading of implicit cognition and we argue that the wider problem may be better understood using the perspective of quantum cognitive theory.

Sometimes it is assumed that dual processes underlie decision-making and agency, one is which is automatic and heuristics-based, and the other is deliberate and rule-based. But this creates the problem of interaction between the two processes and therefore dual process theory only appears to be an *ad hoc* perspective on the problem [2]. Cognitive processing has explicit and implicit functioning that includes a sensory processor, short-term or working memory, and long-term memory (Figure 1) [3],[4]. The sensory processor deals with signals reaching the senses and issues related to focus and intent [5],[6], whereas the actual encoding and retrieval processing, with explicit and implicit functions, is done in the working memory. There is also the question of the capacity of working memory [7], which shows that memory span is inversely related to word length. Subjects remember more short words than long words in recall tests, and the span depends on the number of words the subject can read in about 2 seconds. This speaks to the fact that the executive in the processing is dealing with limits on the computing and memory resources.

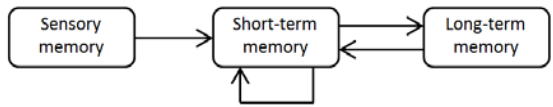

Figure 1. The memory systems

The picture is complicated by the existence of several broader systems of memory (Figure 2) and corresponding cognitions, which include implicit cognition that rests on implicit memory. Although the explicit and implicit memories are usually shown as different blocks, it seems that a better way to see them is through the lens of a single memory system that projects into two or more components based on how the mind interacts with it. This view side-steps the problem of the interaction between different kinds of memories.

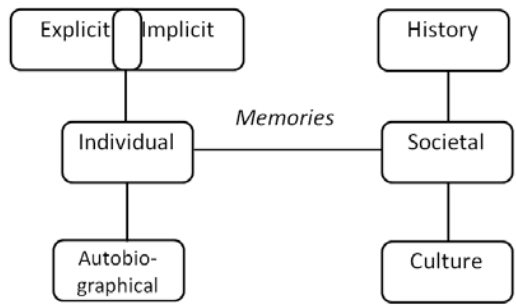

Figure 2. Individual and collective memories

Explicit memory is the conscious storage and recollection of data, and this includes semantic and episodic memory. Since implicit memory represents a set whose contents are not explicit, the model of a set with changing characteristics may be used for it [8]. Implicit memory is associated with priming [9],[10], or the process of a subliminal eliciting of specific responses.

Some aspects to cognition go against expectations associated with rational agents. For example, prospect theory [11][12] describes how humans pick between alternatives with unknown probability that involve risk, where such decisions are based on the potential value of losses and gains rather than the final outcome. In this and other models, one is seeking to find causative relationship between variables in disaggregated or aggregated data, and it is assumed that the agents' decision system is modeled correctly in classical terms. These classical models include connectionist [13],[14] and other biologically-inspired models [15],[16]. Often the peculiarities of the agent behavior are ascribed to complexity and underlying physics and this is taken to include the emergence of consciousness [17].

This apart, it is accepted that quantum processes play a direct or indirect role in biology, as in photosynthesis [18], olfaction, vision [19], bird navigation and other related processes [20],[21],[22]. Although, there is no consensus on models of quantum processing in the brain, there is substantial support for the view that such models are appropriate for cognition [23]. But if applicable only for the processing of information, such models do not represent any fundamental departure from classical processing, excepting in the speed at which processing may take place, since the two are equivalent from a computational theory perspective [24]. On the other hand, if the quantum processing occurs at a higher level of abstraction unrelated to physical processing [25], then there is potential that it will provide new insights into behavior that are not captured by classical models.

There are also proposals that biological memory itself might have a quantum nature. If that were true, it can explain certain peculiarities of human cognition. In certain processing, the decision of the human agent is predicated on beliefs that may be seen as operators that act as filter on the memories or expectations, and belief may also have a quantum basis [26][27]. Quantum languages appear to belong to the language hierarchy associated with the brain [28],[29], so it is likely that processing of sensory and actuator information involves consideration of latent variables with a quantum mechanical basis. Quantum models (e.g. [30]) may be used to explain counter-intuitive data concerning conjunction and disjunction fallacies where real choices sometimes violate the expected utility hypothesis.

In this paper, we examine order and related effects and review several proposals for the use of quantum states and quantum decision trees to understand them. We discuss processing issues concerning implicit cognition which is based on data that the subject is unaware of, and ask if this might have a quantum basis. Lastly, we discuss a maze problem and show how quantum resources provide a solution that is much more efficient than the best classical solution.

**Implicit memory**
Separation between implicit and explicit memory is known to be true for numerous tasks, and it seems to parallel other separations between implicit and explicit knowledge, which in an earlier period was seen through the dichotomy of unconscious and conscious knowledge [31][32][33].

The local and the global aspects of explicit and implicit memory relate to different kinds of perception processes. Hine and Tsushima assert [34] that explicit memory is not influenced by individual perception style but implicit memory is, and they further add: "individual sensory differences have some effects on unconscious processing rather than conscious processing. If this is the case, personal characteristics including perception style would correlate with deep inside the mind, such as the part of iceberg hidden under the sea."

So what is the difference between implicit and explicit memory? Should we see them as multiple memory systems but that appears to be problematic from an underlying philosophical perspective. If we assume that the fundamental memory system is one but it projects its contents based on how the memories are harnessed by the mind, then we stand on better grounds. This is shifting of perspective which allows us to see the differences in our perception to the interaction systems of the central executive and not to the underlying memory system.

In implicit memory, subjects possess a particular kind of knowledge but they are not consciously aware of the knowledge and cannot gain explicit access. Thus patients with various lesions and deficits show implicit knowledge of stimuli that they cannot explicitly perceive, identify, or process semantically. The phenomenon of "blindsight" occurs in patients who claim that they are guessing the location and identity of the visual stimulus but do not "see" anything at all. In alexia without agraphia, a person with injury to the brain may be unable to read but can still write (more on this and other aphasias in [13]). In priming, the exposure to a stimulus influences a response to a subsequent stimulus, and this may be in different domains, such as perceptual, semantic, or conceptual, and this works without conscious guidance or intention.

We don't quite know how individual memories are stored although we know damage to which parts of the brain have specific effect on different kinds of memory [35]. If a memory is associated with a belief, the latter may be a numerical value or a binary variable that represents a choice between polar opposites.

Specific instances of priming may be explained by linked memories, as in the first notes of a song leading to the latter part of it and this can even be explained by recurrent neural network where the entire melody is generated once the initial portion of it is fixed [36].

**Order effects**
Mathematically, if C(A) is the cognition related to a stimulus A, then one can using a preceding stimulus B speak of the cognition C(A|B). If,

$$C(A|B) = f(A|B)$$

$$C(B|A) = f(B|A)$$

Let B be implicit and A be explicit, then we expect to see that

$$C(A|B) \neq C(B|A)$$

If on symmetry grounds, A and B are equally likely and the cognitions are equivalent to probability of recognition, then it follows that

$$f(A,B) \neq f(B,A)$$

As example, consider the extreme case of A and B are two different visual fields for a patient with some damage to the primary visual cortex; let A represent the field where the subject has conscious perception and B represents the field where he does not. The additional processing pathway through the undamaged visual association cortex does not facilitate conscious vision but allows for appropriate subconscious reaction to visual stimuli in this field. Let the subject be presented with one of two images, say apple and banana, in one field and having made a choice he must pick the same image in the other field. The choice in the blind field will help in the corresponding correct decision in the clear field (because of unconscious awareness), whereas the knowledge of the image in the clear field will not alter the decision in the blind field. This is seen most dramatically in the ability to navigate even in the absence of any functional visual cortex. The patient TN, whose visual cortex in both hemispheres was knocked out by successive strokes was able to navigate quite skillfully avoiding many blockages [37].

Such a behavior cannot be true of a classical system if the two choices are independent. If we look at two sets A and B where the elements are chosen at random (Figure 3), then P(AB)=B(BA).

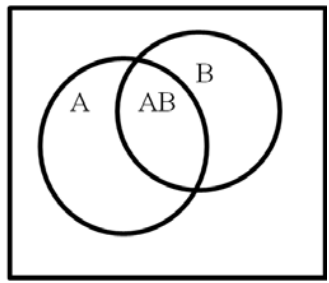

Figure 3. Venn diagram for classical probability where P(AB)=P(BA)

However if the elements are chosen according to further constraints, then this may not be true. For example, the higher order syntactic correlations of English spellings lead to the fact that for the letters "q" and "u", prob(qu) ≠ prob(uq). Such higher order correlations may be viewed as emerging from different branch probabilities in a tree structure. The syntax of the language determines the bigram and n-gram probabilities. Likewise, the syntax associated with memories will determine corresponding order effects, and this might well be the explanation for order effects that have been found in some surveys [38], although there may be other reasons for them also.

There may be non-classical systems [39] that generate unusual kinds of order effects but this requires new assumptions about the nature of memories and their location. On the other hand, even in the absence of no n-gram correlations, unequal order effects are a common characteristic of a quantum system.

Consider the following table from Pew Research [40] that shows typical order effects in surveys and questionnaires.

Table 1. Order effects in surveys

**More Overall Dissatisfaction When Asked After Bush Approval**

| Asked first | Overall satisfaction | % | Bush approval | % |
|---|---|---|---|---|
| | Satisfied | 17 | Approve | 25 |
| | Dissatisfied | 78 | Disapprove | 67 |
| | Don't know | 5 | Don't know | 8 |
| | | 100 | | 100 |
| Asked second | Bush approval | | Overall satisfaction | |
| | Approve | 24 | Satisfied | 9 |
| | Disapprove | 68 | Dissatisfied | 88 |
| | Don't know | 8 | Don't know | 3 |
| | | 100 | | 100 |
| N | | 766 | | 723 |

PEW RESEARCH CENTER Dec. 2008.

**Acquiescence Bias**

**Agree-Disagree Format**
*The best way to ensure peace is through military strength*
(55% agree, 42% disagree)

**Forced Choice Format**
*The best way to ensure peace is through military strength (33%)*
OR
*Diplomacy is the best way to ensure peace (55%)*

PEW RESEARCH CENTER Agree-Disagree question from Oct. 1999. Forced choice question from Sep. 1999.

An analysis of order effects for such examples needs to be situated on stronger methodological grounds for the choices are not quite independent. For example, the variables "satisfied" and "approve" in the first survey are not independent and likewise "military strength" and "diplomacy" need not be independent of each other.

**Order effects related to belief states**

A belief in a decision network may be seen in relation to a small set of variables that directly affect its value in the sense that B is conditionally independent of other variables. As a quantum state associated with memories it need not be taken to be equivalent to the activity in the brain for it may exist in an abstract space. The order effects naturally follow when the variables are quantum states.

The quantum belief state associated with the two-stage experiment may be something like given below (if it is a pure state):

$$|\varphi\rangle = \alpha|AA\rangle + \beta|AB\rangle + \gamma|BA\rangle + \delta|BB\rangle$$

$$|\alpha|^2 + |\beta|^2 + |\gamma|^2 + |\delta|^2 = 1$$

In general, prob$(AB) = |\beta|^2$ is not equal to prob$(BA) = |\gamma|^2$.

*The Disjunction Effect.* Mathematically, the disjunction effect is a counter-intuitive false judgement that the probability P(A or B) is less than either P(A) or P(B).

The motivational setting for the effect is a gamble (with purported 50% chance of win) in which a win is worth $200 and a loss makes one lose $100. If told that they had won the first gamble, 69% of the players decided to play again; told that they had lost, 59% players chose to play again. But only 36% players chose to accept the second gamble if they were not told the outcome of the first gamble [11]. In other words, the decision supports the incorrect P(A) > P(A or B).

If the belief is a quantum state, the reference point may be taken [41] as the superposition state $\sqrt{\frac{2}{3}}|1\rangle + \sqrt{\frac{1}{3}}|2\rangle$ where $|1\rangle$ represents the state of losing $100 is and $|2\rangle$ is the state of winning $200. This balances the two outcomes for the probabilities of the two outcomes are 2/3 and 1/3, respectively, with a utility of $-2/3 \times 100 + 1/3 \times 200 = 0$.

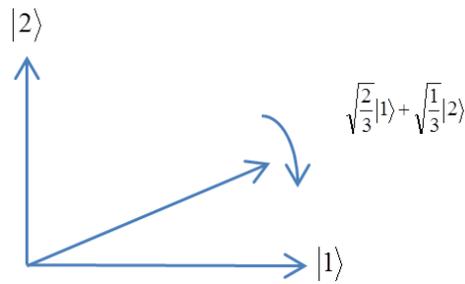

Figure 4. The prospect as a quantum state [41]

The arrow of the reference state in Figure 4 will turn clockwise as the gamble is played successively, without knowing the outcome. On the other hand, the knowledge of the outcome resets the reference at its original location where the utility is zero.

*The Absent-Minded Driver.* In the absent-minded driver the driver must have a strategy to return home choosing from different exit ramps that have varying payoff. With the payoffs of Figure 5, a rational strategy has α=1/3 and a payoff 4/3. But if the driver has access to entangled qubits $\varphi(n) = \frac{1}{\sqrt{2}}(|01\rangle + |10\rangle)$ where "0" represents "don't take exit" and "1" represents "take exit", the payoff is increased to 2.

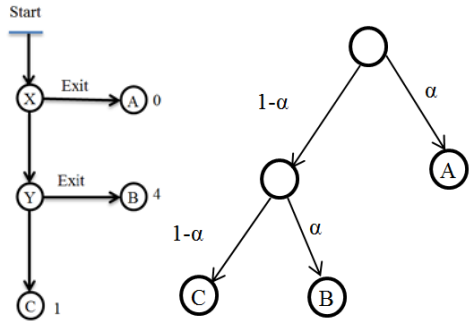

Figure 5. The absent-minded driver and the decision tree [41]

Thus the quantum decision strategy is superior to the best classical strategy and the advantage of quantum processing is due to its ability to have global information. But note that the resources available to the classical and quantum decision agents are different.

**A maze problem**
Imagine that an agent is in a circular room with three doors, two of which return him to the room (with no knowledge of the door that was just used) after passing through a maze, whereas the third one offers him a way out. How many tries will be needed for the person to get out?

If he has only classical resources, the probability function describing his situation is that he will get out on the kth try with a probability of:

$$P(k) = (1-p)^{k-1}p$$

Here p is the probability of using the correct door which is 1/3. He is able to get out on the kth try if he has been unsuccessful on the preceding (k-1) tries. Therefore, the expected number of tries before he gets out is:

$$E(k) = \sum_{k=1}^{\infty} k(1-p)^{k-1}p = \frac{1}{p}$$

In this example, therefore, the person will have to try an average of 3 times before being able to get out.

Now consider the information as a quantum state. Let the state of the qubit on the door be coded in a manner so that 1 represents the correct door and 0 represents the incorrect door. The quantum state has entangled qubits as shown below,

$$\varphi(n) = \frac{1}{\sqrt{2}}(|001\rangle + |110\rangle)$$

The person interacts with the qubit on the door and if it shows 0, he moves to the next door, and stop only when he finds the measurement of the qubit to show 1. There is now a 50% chance that he will choose the first component that will take him to the correct door.

The average number of tries to reach the correct door has now come down to 2, thus implying a fifty percent improvement over the best classical strategy. If the decision protocol is to see which gate has a bit that is different from others, then the quantum resource will reveal the gate right away.

If the person is only told that a 1 corresponds to a productive gate and not how many gates take him out, then even in this reduced information situation, the person will be able to get out in 2 tries.

**Conclusions**

This paper examined some common assumptions regarding the decision-making internal environment for intelligent agents. Since it is clear that such decisions include both global and local information, we investigated if the perspective of quantum cognition can help obtain better understanding of the decision process. In specific, we considered implicit cognition and explored if certain inaccessible may be best modeled as quantum states. Some problems with intelligent agents were reviewed and cases where quantum decision theory provides superior performance were investigated. A problem involving a maze network was considered and comparisons made between classical and quantum decision scenarios for it.

The perspective of quantum states appears to provide new insight into the workings of agents in different decision environments. The belief vector is a superposition vector of mutually exclusive outcomes which rotates when it interacts with the vector of internal and external states. This may not provide us a method of predicting behavior, but it can be useful for its understanding. Furthermore, it may be useful in analysis of order effects in healthcare systems as in drug effectiveness and provide insight into the placebo effect [42].

Most significantly, the paper argues for careful design of questionnaires and surveys that clearly rule out classical explanations. Doing so will provide a conceptual breakthrough that will illuminate seemingly counter-intuitive behavior of intelligent agents.